\documentclass[11pt]{article} %
\usepackage{wrapfig2}

\usepackage[preprint]{acl}

\usepackage{times}
\usepackage{latexsym}
\usepackage{placeins}

\usepackage[T1]{fontenc}

\usepackage[utf8]{inputenc}

\usepackage{microtype}

\usepackage{inconsolata}

\usepackage{amsmath,amsfonts,bm}

\newcommand{\prompt}{\theta}

\newcommand{\promptOutput}{f(x_i; \theta)}
\newcommand{\promptOutputNewWeight}{f(x_i; \theta')}

\newcommand{\promptOutputShort}{\hat{y}}

\newcommand{\evaluation}{\mathcal{L}(x_i, y_i, \promptOutput)}
\newcommand{\evaluationNewWeight}{\mathcal{L}(x_i, y_i, \promptOutputNewWeight)}
\newcommand{\evaluationShort}{\mathcal{L}(x_i, y_i, \promptOutputShort)}

\newcommand{\evaluationIncorr}{\mathcal{L}(x_i, \tilde{y}_i, \promptOutputShort)}

\newcommand{\feedbackForOutput}{\frac{\partial \evaluationShort}{\partial \promptOutputShort}}

\newcommand{\feedbackForPrompt}{\frac{\partial \evaluationShort}{\partial \prompt}}

\def\eqref#1{equation~\ref{#1}}

\def\1{\bm{1}}

\DeclareMathAlphabet{\mathsfit}{\encodingdefault}{\sfdefault}{m}{sl}
\SetMathAlphabet{\mathsfit}{bold}{\encodingdefault}{\sfdefault}{bx}{n}

\usepackage{hyperref}
\usepackage{url}
\usepackage{graphicx}
\usepackage{subfig}
\usepackage{csquotes}
\usepackage{markdown}
\MakeOuterQuote{"}

\usepackage{amsthm}
\newtheorem{definition}{Definition}

\title{Textual Gradients are a Flawed Metaphor for\\ Automatic Prompt Optimization}

\author{Daniel Melcer\thanks{Work performed during an internship at Amazon} \\ Northeastern University \\ Boston, MA, USA \\ \texttt{daniel@melcer.dev}
\And Qi Chen \and Wen-Hao Chiang  \\ {\bf Shweta Garg \and Pranav Garg} \\ {\bf Christian Bock}
 \\ AWS AI Labs \\ New York, NY, USA \and Seattle, WA, USA \\ \texttt{\{qic,cwenhao,shwegarg,}\\\texttt{prangarg,bocchris\}@amazon.com}}

\newcommand{\hatch}[1]{\raisebox{-1.7pt}{\includegraphics[height=1.3\fontcharht\font`\B,page=#1]{images/rectangles.pdf}}}

\begin{document}

\maketitle

\begin{abstract}
A well-engineered prompt can increase the performance of large language models; automatic prompt optimization techniques aim to increase performance without requiring human effort to tune the prompts.
One leading class of prompt optimization techniques introduces the analogy of textual gradients.
We investigate the behavior of these textual gradient methods through a series of experiments and case studies.
While such methods often result in a performance improvement, our experiments suggest that the gradient analogy does not accurately explain their behavior.
Our insights may inform the selection of prompt optimization strategies, and development of new approaches.
\end{abstract}

\section{Introduction}

Large Language Models (LLMs) have grown significantly in capability and popularity in recent years. 
Tasks that were once considered out-of-reach, such as graduate-level question answering \citep{rein2023gpqagraduatelevelgoogleproofqa}, top-level competitive math \citep{Castelvecchi_2025}, and advanced automated software engineering \citep{wang2025openhandsopenplatformai}, are now possible with the latest generation of frontier models.

Despite their increase in power and generality, they are not perfect.
While fine-tuning can be a useful tool to specialize a model for a specific task, it is generally expensive, and often requires a large amount of data \citep{parthasarathy2024ultimateguidefinetuningllms}.
Furthermore, many of the best-performing models are only available behind an API---their weights are not available, and fine-tuning may be impossible.

In many cases, a well-engineered prompt or sequence of prompts enables an increase in task-specific performance, without requiring access to the model weights \citep{mishra2022reframinginstructionalpromptsgptks}.
However, constructing such a prompt may be a difficult process, relying on a combination of intuition and trial-and-error.
Several model providers have released prompt engineering guides \citep{openai2025promptengineering, anthropic2025promptengineering}, but prompt construction with these guides remains an art.

\emph{Automatic Prompt Optimization} (APO) methods aim to automate the process of creating such a well-engineered prompt \citep{zhou2023largelanguagemodelshumanlevel,yang2024largelanguagemodelsoptimizers,fernando2023promptbreederselfreferentialselfimprovementprompt,guo2025evopromptconnectingllmsevolutionary}.
A recent development---\emph{textual-gradient} based methods \citep{pryzant2023automaticpromptoptimizationgradient,cheng2024traceautodiffgenerativeoptimization,yuksekgonul2024textgrad}---liken a model's system prompt to a learnable parameter.
In a traditional machine learning system, the gradient of a model's parameters describes how the parameters should change to achieve an objective.
With the gradient analogy, feedback about how to improve the system prompt is treated as the prompt's ``gradient''.
A prompt engineer could implement a familiar training loop, perform ``gradient descent'' on the prompt, and obtain an ``optimized'' prompt after some number of training iterations.

While this method usually obtains a well-performing prompt, our experiments show that the gradient analogy does not accurately explain the behavior of APO.
In Section \ref{section:overview}, we provide an overview of the gradient hypothesis.
In Section \ref{section:textual-gradient-apo-can-help}, we perform an ablation study where we would expect a significant decrease in performance under the gradient hypothesis, but generally do not observe such a decrease.
Finally, in Section \ref{section:case-studies}, we introduce several case studies to further explore the behavior of automatic prompt optimization.

\section{Related Work}

With access to a model's weights, it is possible to perform backpropagation to a prompt's embedding space to learn a task-specific prompt \citep{shin2020autopromptelicitingknowledgelanguage}.
However, with closed models, gradient-free methods are necessary for prompt optimization.

\citet{jiang2020knowlanguagemodelsknow} explored how backtranslation can be used to generate variations of a prompt.
\citet{prasad2023gripsgradientfreeeditbasedinstruction} instead use a phrase-substitution heuristic to generate variants.

As LLMs have grown more powerful, more recent APO methods use the LLM itself for prompt generation.
\citet{zhou2023largelanguagemodelshumanlevel} and \citet{yang2024largelanguagemodelsoptimizers} use an LLM to generate a new prompt based on a scoring model and evaluation accuracy, respectively, while
and \citet{fernando2023promptbreederselfreferentialselfimprovementprompt} and \citet{guo2025evopromptconnectingllmsevolutionary} use an LLM to implement mutation and crossover operators for an evolutionary algorithm.
\citet{do2024promptoptimizationadversarialincontext} uses an LLM to mutate prompts as part of an adversarial learning process, while \citet{gupta2024metareflectionlearninginstructionslanguage} uses a self-reflection process to learn specific insights that are concatenated to the prompt.
\citet{khattab2023dspycompilingdeclarativelanguage} released a package that allows for automatically learning prompts in a pipeline, where an LLM is invoked at several specialized stages to achieve a complex goal. 

\citet{pryzant2023automaticpromptoptimizationgradient} introduces a gradient-like analogy for generating new prompts from evaluation feedback; several works expand this analogy and provide an autodiff-like implementation \citep{cheng2024traceautodiffgenerativeoptimization, yuksekgonul2024textgrad}.
\citet{wan2024teachbettersmarterinstructions} compares the performance of several prompt optimization methods in combination with methods for selecting in-context examples.

\citet{pan-etal-2023-context} and \citet{min-etal-2022-rethinking} show that the performance of prompts with incorrect or random in-context examples is often comparable to the performance with correct in-context examples.

In some agentic systems, an LLM is used to tune the optimization process \citep{zelikman2024selftaughtoptimizerstoprecursively}, or even modify the code of the agent itself \citep{zhang2025darwingodelmachineopenended,robeyns2025selfimprovingcodingagent,yin2025godelagentselfreferentialagent}.

These are only a sample of automatic prompt optimization methods; we refer readers to \citet{ramnath2025systematicsurveyautomaticprompt} for a broader overview of APO.

\section{Overview}

\label{section:overview}

Any APO method requires some mechanism to generate a prompt.
We first describe two broad categories of this mechanism: prompt-only, and feedback-driven generation. 
There are additional categories that we do not address here, such as evolutionary algorithms \citep{fernando2023promptbreederselfreferentialselfimprovementprompt}, or hybrid mechanisms \citep{pryzant2023automaticpromptoptimizationgradient}.

In \textbf{prompt-only} generation,\footnote{Prompt-only generation is sometimes referred to as Monte-Carlo generation, when used to generate many prompts with a later selection step. We instead distinguish the prompt generation method (prompt-only or feedback-driven) from the selection method (validated with $n$ variants, or naive).} only the existing prompt is used to generate a new prompt. 
Early implementations use a variety of methods to generate variants of a given prompt, such as phrase substitution \citep{prasad2023gripsgradientfreeeditbasedinstruction} or backtranslation \citep{jiang2020knowlanguagemodelsknow,xu2022gpsgeneticpromptsearch}.
Recent versions of prompt-only generation typically provide an LLM with instructions to rephrase the prompt without changing its meaning \citep{do2024promptoptimizationadversarialincontext}.

With \textbf{feedback-driven} prompt generation, the prompt is used to generate one or more outputs, and then feedback based on these outputs is used as part of the context for generating a new prompt.

\subsection{The Gradient Hypothesis}
\label{subsec:gradient-hypothesis}

A common variant of feedback-driven generation is the \textbf{gradient-like} method \citep{pryzant2023automaticpromptoptimizationgradient}, which relies on the following assumption:

\begin{definition}[Gradient Hypothesis]
    Textual feedback from an LLM behaves like a gradient, and it is possible to apply the chain rule to this feedback.
\end{definition}

We illustrate the gradient analogy with an analog of the standard training loop.

Let $f : V^* \times V^* \rightarrow V^*$ be a pretrained LLM with token space $V$ and frozen weights; it is parameterized by system prompt $\prompt \in V^*$.
Given a dataset $D$, and question-answer pair $(x_i, y_i) \in D$, the output of the LLM is $\promptOutputShort = \promptOutput$.

In lieu of a numeric loss function such as mean-square error, a textual loss function may be a template substitution such as $\evaluationShort = $ ``Question: $x_i$. Provided answer: $\promptOutputShort$, Correct answer: $y_i$.''
Alternatively, $\evaluationShort$ may itself invoke an LLM.

According to the gradient hypothesis, asking an LLM ``How can the answer ($\promptOutputShort$) be changed to improve its evaluation ($\evaluationShort$)?'' is analogous to computing $\feedbackForOutput$.

Similarly, asking the LLM ``How can the system prompt ($\prompt$) be changed such that the resulting output ($\promptOutputShort$) improves according to its feedback $\left(\feedbackForOutput\right)$?'' applies the chain rule to obtain $\feedbackForPrompt = \frac{\partial \promptOutputShort}{\partial \prompt} \feedbackForOutput$.

Finally, the LLM is directed to ``Update the prompt ($\prompt$) using its feedback $\left(\feedbackForPrompt\right)$.''
This is analogous to taking an optimizer step to obtain new prompt $\prompt' = \prompt - \alpha\feedbackForPrompt$.\footnote{The ``step size'' $\alpha$ is undefined in this analogy, but \citet{yuksekgonul2024textgrad} introduce a momentum-like behavior.}
According to the gradient hypothesis, $\evaluationNewWeight < \evaluation$ for some definition of ``$<$''.

In this paper, our usage of the gradient analogy is limited to these three operations, and addition (implemented as string concatenation).
However, packages such as those introduced in \citet{yuksekgonul2024textgrad,cheng2024traceautodiffgenerativeoptimization} implement full ``automatic differentiation'' with textual gradients, using a PyTorch-like syntax.

With traditional gradient descent, the new parameters $\theta'$ are immediately used for future iterations (\textbf{naive} selection).
However, many APO methods include an extra step---they generate one or more prompts $\prompt'_1, \ldots, \prompt'_n$ using the above process, and then test each of these prompts on a validation set.
The best prompt is chosen, though only if it improves on the previous prompt's ($\prompt$) performance.
We term this \textbf{validated} prompt selection.

\section{Analysis of Gradient-like APO}

\label{section:textual-gradient-apo-can-help}

\subsection{Ablation Study}

\label{subsection:textual-gradient-vs-mc-onestep}

We introduce the following research questions to test the gradient hypothesis.

\textbf{RQ1: Does the gradient-like structure matter?} Rather than using a series of three prompts to implement the gradient analogy from Section \ref{subsec:gradient-hypothesis}, we implement a \textbf{One-step} variant generation routine that asks the LLM to directly generate a new prompt $\prompt'$ using evaluation $\evaluationShort$.

\textbf{RQ2: Does a correct loss function matter?} In the \textbf{Incorrect Evaluation} configurations, we simulate an incorrect loss function by deliberately inserting an incorrect ground-truth answer $\tilde{y}_i$ in the evaluation $\evaluationIncorr$. In the \textbf{No Evaluation} configuration, the template doesn't include any ground-truth answer $y_i$, correct or incorrect.

\textbf{RQ3: Does validation matter?} We run both \textbf{naive} and \textbf{validated} versions of experiments.

\textbf{RQ4: Does using a training dataset matter?}
We implement \textbf{Prompt-Only} optimization, which does not use training data at all.
In addition to the standard \textbf{Rewrite} variant (ask an LLM to rephrase the prompt without changing its meaning), we also try asking an LLM to improve the prompt for quality and correctness (the \textbf{Improve} variant).

All prompts are included in Appendix \ref{appendix:prompts}.

\begin{figure*}[t]
    \centering
    \includegraphics[width=1.0\linewidth]{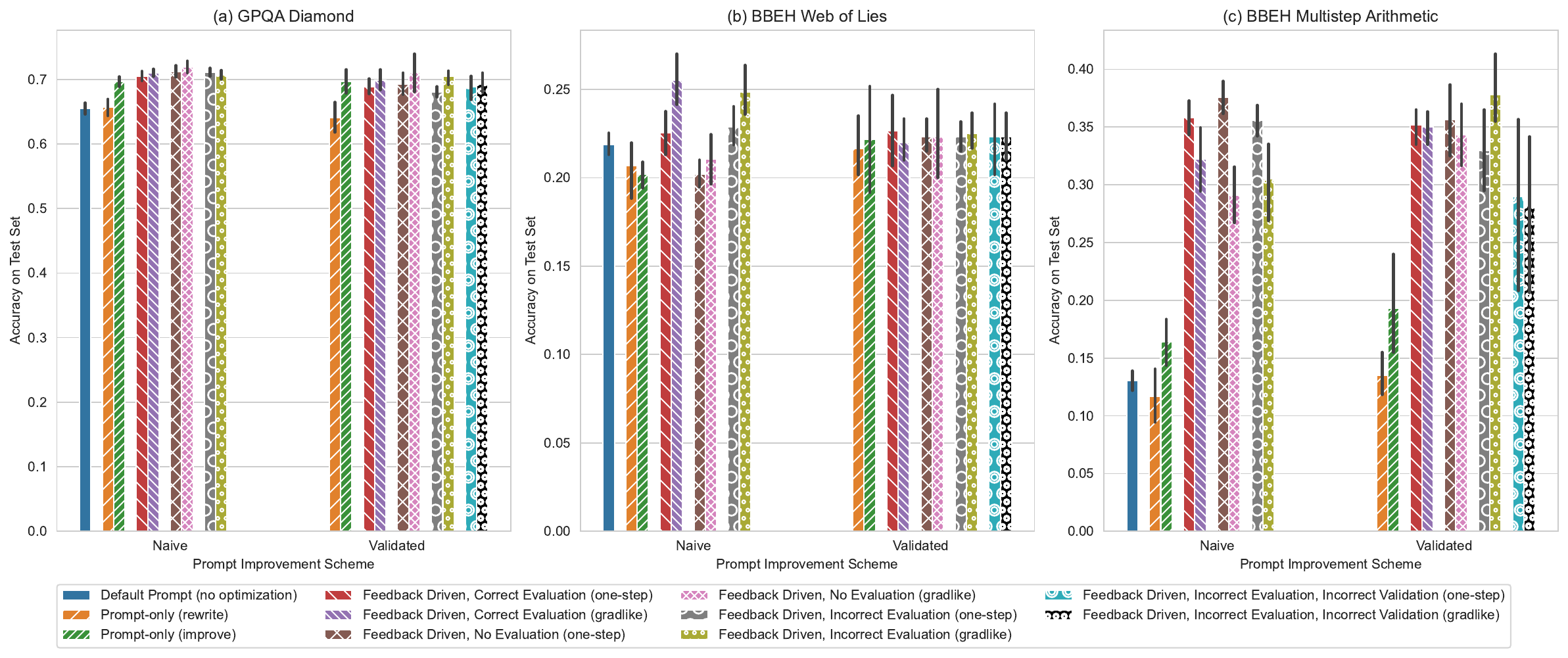}
    \caption{Performance on (a) GPQA Diamond (b) Web of Lies (c) Multistep Arithmetic, all with 95\% CI. 30 trials for each naive configuration; 5 for each validated.}
    \label{fig:feedback-schemes}
\end{figure*}

We use three datasets.
Our first dataset is \textbf{GPQA Diamond}\footnote{GPQA licensed with CC-BY 4.0; BBEH with Apache 2.0} \citep{rein2023gpqagraduatelevelgoogleproofqa}, containing 198 English-language expert-level science questions. For training and validation, we use 30 and 50 questions, respectively, that are from in the broader GPQA dataset, rather than the ``diamond'' subset.

Our other two datasets are subsets of \textbf{BigBenchExtraHard} \citep{kazemi2025bigbenchextrahard}. 
Each contains 200 problems; we reserve 30 problems for training and 50 for validation, leaving 120 for testing.
\textbf{Web of Lies} is a generalization of the classic logic puzzle format: ``Alice says that Bob is lying; Bob says that Charlie is telling the truth; Charlie is lying. Is Alice telling the truth?''
The model is asked to find the truth value of three characters; it is also possible for the truth-telling status of a character to be indeterminate. 
\textbf{Multistep Arithmetic} requires the language model to reason about custom math operators. 

For the incorrect-evaluation trials, we obtain $\tilde{y}_i$ by selecting the answer one index greater than $y_i$ in GPQA, creating a random answer in Web of Lies, or adding a random integer between -10 to 10 to the true $y_i$ in Multistep Arithmetic.
The question text is used as the random seed; the same question will always have the same $\tilde{y}_i$. 

For all experiments, we begin with a default prompt such as ``Answer the given multiple choice question.''
The prompt is appended to formatting instructions that are constant throughout learning; for example, writing the answer inside specific tags.

We run 30 trials for each naive configuration, and 5 trials for each validated configuration, generating 3 prompt variations at each batch during validated trials.
All configurations use Claude 3.7 Sonnet, with temperature $0.5$, top-p sampling with $p=0.95$, and a 5000-token limit.
We run 10 iterations of the training loop (1 epoch), with a batch size of 3.
Total token usage is reported in Appendix \ref{appendix:token-usage}.

\subsubsection{Results \& Discussion}

\label{subsubsection:main-apo-results}

The test set accuracies are presented in Figure \ref{fig:feedback-schemes}.

\textbf{RQ1: The effect of the gradient-like structure is dataset-dependent.}
Gradient-like APO consistently led to improved performance compared to an un-optimized prompt (Figure \ref{fig:feedback-schemes}, \hatch{5} vs. \hatch{1}).
However, gradient-like APO does not consistently outperform one-step APO (Figure \ref{fig:feedback-schemes}, adjacent pairs of gradient-like vs. one-step, \hatch{5} vs. \hatch{4}).

We use a two-sided permutation t-test to further compare the performance of gradient-like versus one-step feedback-driven APO, for both naive and validated correct-evaluation configurations.
As this experiment involves two comparisons for each dataset, we apply the Bonferroni correction \citep{Vickerstaff_Omar_Ambler_2019} to obtain a significance threshold of $0.025$.
We find that in Web of Lies, gradient-like APO outperforms one-step feedback-driven APO in the naive setting ($p=0.0037$). 
We did not find a significant performance difference for other configurations or datasets; we cautiously observe that this trend may be reversed in the naive configuration of Multistep Arithmetic, though this result does not reach the significance threshold ($p=0.030$).
Full p-values are listed in Appendix \ref{appendix:p-values}. 

Under the assumption that the evaluation accuracy of prompt optimization is normally distributed, with a large effect size ($d=0.8$), the statistical power of this test is $0.79$ for the naive trials, and $0.12$ for the validated trials. 
The disjunctive statistical power is $0.81$.
Except as otherwise stated, we use these assumptions for calculating the statistical power for the remainder of the paper.

\textbf{RQ2: An incorrect evaluation function does not usually hurt performance, but a missing evaluation function may.}
The gradient hypothesis implies that using an incorrect training set would lead to significantly worse test performance than with a correct training set.
However, we do not observe this behavior.
Visually, this is illustrated in Figure \ref{fig:feedback-schemes} by comparing the correct evaluation trials (\hatch{4}\hatch{5}) to the incorrect evaluation trials (\hatch{8}\hatch{9}).

We use a one-sided permutation t-test to further analyze these comparisons.
There are 4 comparisons for each dataset (naive and validated selection; and one-step and gradient-like generation).
Applying the Bonferroni correction, we obtain a significance threshold of $0.0125$.
The statistical power for each naive trial is $0.79$; $0.12$ for each validated trial.\footnote{While the corrected $p$ threshold is lower than with RQ1, we use a one-sided test, resulting in the same per-test power.}
The disjunctive power is $0.96$.
We do not find any significant decrease in accuracy when training using incorrect evaluation labels.

We perform the same analysis, comparing against training with missing evaluation labels (\hatch{4}\hatch{5} vs. \hatch{6}\hatch{7}) rather than with incorrect labels.
We found that in the naive trials for Web of Lies, there is a significant decrease in performance: $p=0.0022$ and $p < 0.0001$ for one-step and gradient-like trials respectively.
However, we did not observe such a decrease in any other datasets.
The power analysis is the same as with the incorrect evaluation.

We believe that the most likely explanation for these observations is that APO usually assists with the general format of the task, or with meta-instructions---such as to incorporate a chain-of-thought style when answering, avoid concluding ``unknown'', or to answer concisely to avoid hitting a length limit---but it rarely incorporates insights from any one particular question.
This is explored further in Section \ref{subsection:bbehwol-case-study}.

\textbf{RQ3: Validation has inconsistent effectiveness.}
For each dataset, we compare all eight naive configurations to the corresponding validated configuration.
In Figure \ref{fig:feedback-schemes}, this is the comparison between all bars with a ``Naive'' label against the corresponding bar with a ``Validated'' label.

We use a one-sided permutation t-test with the corrected $p$ threshold of $0.00625$.
The test for each configuration has a power of $0.18$, resulting in a disjunctive power of $0.79$.
We do not find a significant improvement with validated prompt selection.

We also run an exploratory test to determine if an incorrect validation set hurts performance, in cases where an incorrect training set is used with validated prompt selection (\hatch{8}\hatch{9} vs. \hatch{10}\hatch{11}).
There are two comparisons for each dataset; using a one-sided permutation t-test with a significance threshold of $0.025$, we find that incorrect validation data hurts performance in Multistep Arithmetic, with gradient-like training ($p=0.0238$). 
However, because we only run 5 validated trials per configuration due to their high token usage, this test requires an effect size of $d=1.51$ to reach a disjunctive power of $0.8$.
Additional trials are required to rule out smaller effect sizes for the other two datasets.

\textbf{RQ4: Prompt-only optimization sometimes works, but feedback-driven is often better.}
For each dataset, we compare the effectiveness of ``rewrite'' prompt-only generation, in naive and validated configurations.
As shown in Figure \ref{fig:feedback-schemes}, the ``rewrite'' prompt-only configuration (\hatch{2}) performs similarly to the un-optimized prompt (\hatch{1}) configuration.
Using a one-sided permutation t-test, with 2 comparisons for each dataset (corrected significance threshold of $0.025$), we do not find a significant improvement over the default.
The per-configuration power is $0.86$ for the naive experiment and $0.36$ for the validated experiment, providing a per-dataset disjunctive power of $0.91$.

We repeat the analysis for prompt-only ``improve'' APO (\hatch{3} vs. \hatch{1}), and find an improvement in some datasets---both GPQA ($p\approx10^{-5}$ for naive, $p=0.0008$ validated) and Multistep Arithmetic ($p=0.0025$ for naive, $p=0.0004$ validated).

While a more directed prompt may allow prompt-only generation to outperform the default prompt, feedback-driven APO is usually better.
We compare prompt-only generation to gradient-like APO (\hatch{3} vs. \hatch{5}) for both naive and validated experiments ($p$ threshold of $0.025$, naive power $0.86$, validated power $0.20$, disjunctive power $0.89$) and find that gradient-like APO outperforms prompt-only ``improve'' APO in Web of Lies using naive prompt selection ($p \approx 10^{-5}$), and Multistep Arithmetic for both naive and validated selection ($p \approx 10^{-5}$ and $0.0040$, respectively).

\subsection{Direct vs. Critic-based Evaluation}

\begin{figure*}[t]
    \centering
    \includegraphics[width=\linewidth]{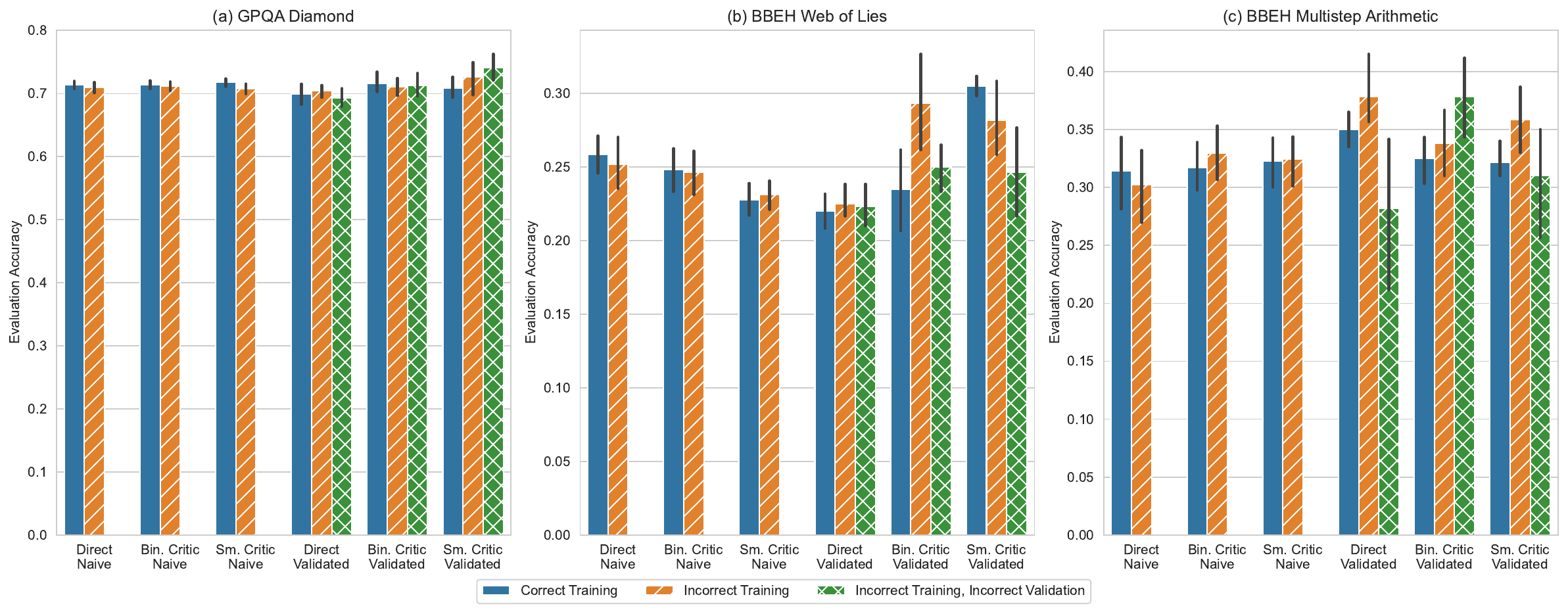}
    \caption{Performance of direct and critic-based prompt optimization; naive and validated variants, using both a correct and incorrect validation set. 30 trials in naive configurations; 5 trials in validated configurations; 95\% CI.}
    \label{fig:additional-incorrect-training-variants}
\end{figure*}

\label{subsection:variants-of-gradlike}

We analyze the case where LLM-generated feedback is used (\textbf{Critic-Based Evaluation}), rather than a template based on the actual answer (\textbf{Direct Evaluation}).
This differs from the ungrounded \emph{No Evaluation} configuration in Section \ref{subsection:textual-gradient-vs-mc-onestep}; with critic-based evaluation, the critic's prompt is itself optimized based on the accuracy of its outputs.

Critic-based evaluation mirrors the structure of two adversarial networks \citep{goodfellow2014generativeadversarialnetworks}. It was first instantiated in the textual domain by \citet{do2024promptoptimizationadversarialincontext}, albeit with prompt-only optimization. 
We use a variant of this framework to optimize the generator and critic prompts simultaneously:

An LLM critic, parameterized by system prompt $\phi$, is used to generate evaluation $\mathcal{L}_G(x_i, \promptOutputShort; \phi)$.
This evaluation is used by the standard APO framework to update the generator prompt $\prompt$.
Meta-evaluation $\mathcal{L}_C(\evaluationShort, \mathcal{L}_G(x_i, \promptOutputShort; \phi))$ is created for the critic by comparing its output against a direct evaluation of the generator's output. An independent instance of APO is used to obtain new critic prompt $\phi'$ using this evaluation, using the meta-prompts in Appendix \ref{appendix:prompts-indirect}.

We implement both \textbf{Binary Critic-Based} evaluation, where the critic is instructed to provide a binary judgment of the generator's output, and \textbf{Smooth Critic-Based} evaluation, where the critic is instructed to assign a score between 0 and 100.

We run 30 trials for all naive variants and 5 for validated variants; results are shown in Figure \ref{fig:additional-incorrect-training-variants}.

As shown by the adjacent correct vs. incorrect trials for the new variants (Figure \ref{fig:additional-incorrect-training-variants}, \hatch{1} vs. \hatch{2}), an incorrect label (replacing $y_i$ with $\tilde{y}_i$) continues to not harm performance in the critic-based variant.
We repeat a one-sided permutation t-test in these additional variants; with 4 new variants per dataset (smooth and binary; naive and validated), we use the Bonferroni correction to obtain a $p$ threshold of $0.0125$.
The power analysis is identical to that in Section \ref{subsubsection:main-apo-results}, RQ2.
We again do not find a significant decrease in performance, inconsistent with a gradient-like interpretation of textual feedback.

We also analyze whether critic-based training performs differently from the equivalent direct configuration
This is visually illustrated in Figure \ref{fig:additional-incorrect-training-variants} as the comparison between critic-based trials and their adjacent direct configurations.

We statistically test this with a two-sided permutation t-test.
There are 8 comparisons per dataset, for a corrected $p$ threshold of $0.00625$.
In Web of Lies, smooth critic-based training results in a lower performance than direct training in the naive case ($p=0.0010$).
However, this trend may be reversed in the validated case---we cautiously observe that several independent smooth critic-based validated trials in Web of Lies have results with $p$-values slightly above the significance threshold ($p=0.0079$ with correct training; $p=0.0159$ with incorrect training).
As the results are likely correlated with each other, the Bonferroni correction may be too conservative in this instance.

We investigate the best-performing trials in Section \ref{subsection:bbehwol-case-study} to better understand the circumstances that lead to improved performance in those trials.

We also test if validated configurations perform better than their equivalent naive counterparts.
There are four new comparisons per dataset, so we use a corrected $p$ threshold of $0.0125$.
In Web of Lies, we find a significant increase in performance with validated training for smooth critic-based feedback ($p < 0.0001$ with correct evaluations, $p=0.0008$ with incorrect evaluation).
We do not find a significant improvement with validation in the other datasets.
As there are relatively few validated configurations, an effect size of $d=0.91$ is required to reach a power of $0.8$; more data is required to reliably detect lower effect sizes.

Finally, we run another exploratory test to assess if incorrect validation data hurts performance, compared to training with correct validation data and incorrect training data (Figure \ref{fig:additional-incorrect-training-variants}, \hatch{2} vs. \hatch{12} in validated configurations).
Using a one-sided permutation t-test (significance threshold of $0.025$), we do not find any statistically significant decrease in performance with incorrect validation data.
The power analysis is identical to the final exploratory test in Section \ref{subsubsection:main-apo-results}, RQ3.

\subsection{(Not) Overfitting the Training Data}

\label{subsection:overfit}

\begin{figure*}[t]
    \centering
    \includegraphics[width=0.9\linewidth]{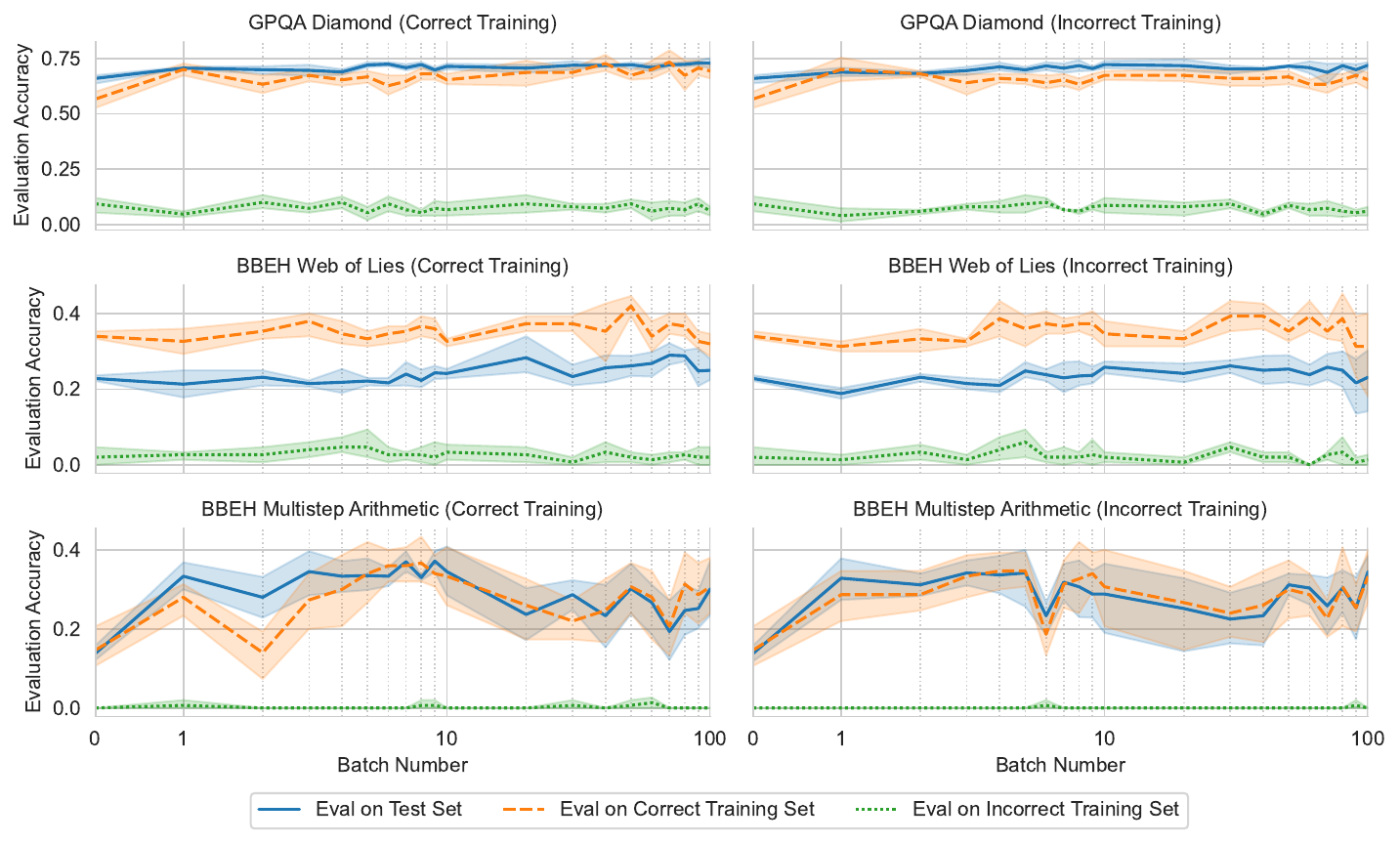}
    \caption{Performance on training set, evaluation set, and an incorrect training set, over 100 iterations (10 epochs). Due to the small training set and uneven distribution of difficult problems, the training set for Web of Lies is somewhat easier than the test set, while the opposite is true for GPQA. 5 trials in all configurations; 95\% CI.}
    \label{fig:overfit}
\end{figure*}

In a traditional gradient-descent based system, training for many epochs on a small dataset will result in overfitting, where performance on the training set continues increasing, but performance on the test set flattens or decreases.
If textual gradients behaved as traditional gradients, we would expect similar overfitting behavior.
We evaluate potential overfitting by training for 100 steps (10 epochs, batch size 3, training set size 30).

We run 5 trials of APO for each dataset.
As shown in Figure \ref{fig:overfit} (left), there is no evidence of gradient-like APO overfitting the training data.
The performance on the training data after 10 epochs is similar to the performance after the first epoch.

Another behavior that we would expect under the gradient hypothesis is that after training on data with a consistently incorrect label, the model should learn this incorrect data.
As Figure \ref{fig:overfit} (right) demonstrates, textual gradient descent doesn't exhibit this behavior, failing to learn the training data.

\section{Case Studies}
\label{section:case-studies}

We first investigate a particularly performant prompt in the Web of Lies domain, to understand the factors that lead to high accuracy when using this prompt.
We then further investigate the behavior of validated prompt selection by replicating selected outliers from Section \ref{section:textual-gradient-apo-can-help}.

\subsection{Right Feedback, Wrong Reason}
\label{subsection:bbehwol-case-study}

We selected one prompt that performed particularly well in Web of Lies, from a validated smooth critic-based run, and analyzed both why this prompt performed so well, as well as the evolution of this prompt. 
An excerpt from this prompt is as follows; it is reproduced in full in Appendix \ref{appendix:case-study-text}:

\begin{displayquote}
\small
When answering logic puzzles...

1. Begin by systematically organizing all relevant information:
[8 lines skipped]

2.  Develop a systematic and exhaustive approach for truth-teller/liar puzzles:
[19 lines skipped]

3. Be EXTREMELY persistent and thorough - NEVER accept ``unknown'' prematurely:

   - CRITICAL: These puzzles are ALWAYS designed to have definitive answers - if you're tempted to conclude ``unknown,'' this is a STRONG SIGNAL that you need to dig deeper

[Prompt continues for 49 additional lines]
\end{displayquote}

\begin{figure}[ht]
    \centering
    \subfloat[Including every section except $X$.]{
                \includegraphics[width=0.9\linewidth]{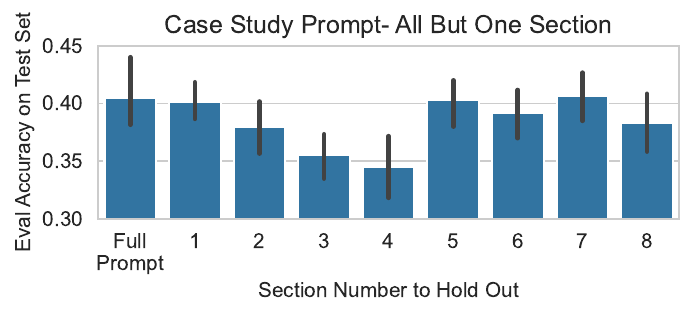}
    }
    \\
    \subfloat[Prompt only contains section $X$.]{
                \includegraphics[width=0.9\linewidth]{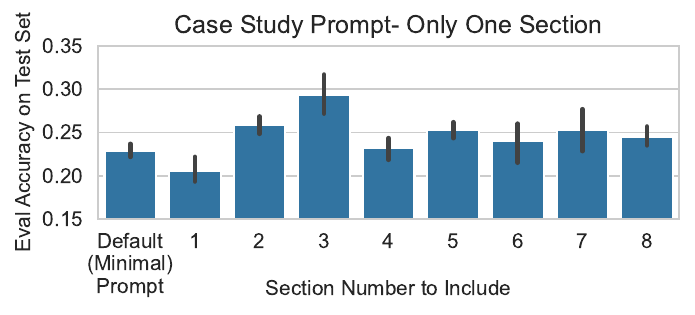}
    }
    \caption{\label{fig:case-study-by-section}Evaluation accuracy for ablations of the case study prompt (Appendix \ref{appendix:case-study-text}); 5 trials, 95\% CI.}
\end{figure}

The prompt is broken up into eight substantive sections.
To determine the most important part of the prompt, we evaluated the performance of each section of this prompt.
We run ablations that include only one section of the prompt, or hold out one section of the prompt, and evaluate the resulting prompt on the test set.
We evaluated each ablation five times.
As shown in Figure \ref{fig:case-study-by-section}, the third section has an impact on the evaluation accuracy.

In Web of Lies, it is possible for the answer to be ``unknown'', in cases where it is truly impossible to deduce the truth-telling status of one of the characters.
However, this is uncommon; usually, when the LLM answers ``unknown'', it is wrong.
The third section of the prompt contains a form of \textit{prevalence hacking}: a strongly worded instruction to never predict the minority class (answering ``unknown'') to increase performance on the majority classes (definitive yes/no answers).

We traced the creation of this section.
In the batch where this section was first created, the generator outputted ``unknown'' for all answers.
As this was a critic-based evaluation trial, the LLM critic rated the generator's output as incorrect, producing the following evaluation ($\evaluationShort$):

\begin{displayquote}
\small
The correct answer is that Raymond, Murphy, and Sima all tell the truth (yes, yes, yes). This can be determined through careful analysis... 

Since the final answer provided is incorrect (unknown, unknown, unknown), the score must be 0, regardless of the partial progress made...
\end{displayquote}

Note that, for this question, the LLM-based critic provided incorrect information! 
One person's truth-telling status is truly indeterminate, but with a critic-based evaluation, the ground truth can only be used to improve the critic's prompt; the critic's output is used to improve the generator's prompt.

The output feedback $\left(\feedbackForOutput\right)$ correctly notes that the generator is too likely to output unknown---but it goes further and declares that an ``unknown'' should never occur:

\begin{displayquote}
\small
2. **Premature declaration of ``unknown''**: All three solutions gave up too easily...

Remember that these logic puzzles are designed to have definitive answers, so if the solution seems incomplete, that's a sign to dig deeper...
\end{displayquote}

The feedback about improving the generator $\left(\feedbackForPrompt\right)$ emphasizes such a declaration:

\begin{displayquote}
\small
    1. **Emphasize persistence and thoroughness**: The prompt should strongly emphasize that these puzzles are designed to have definitive answers...never give up and declare ``unknown'' until truly exhausting all logical possibilities...
\end{displayquote}

Thus, the prompt is updated with incorrect instructions that improve its performance.

In contrast, when direct evaluation is used---using a template $\evaluationShort$ instead of an LLM critic---the feedback must account for the fact that ``unknown'' is sometimes the right answer. 
Typically, the prompt will learn softer instructions, such as to try multiple approaches before concluding unknown.
While this is more factually correct compared to a prompt that effectively bans concluding ``unknown,'' the LLM will often incompletely follow these instructions, give up early, and answer ``unknown'', reducing the test performance.

\subsection{Validation: Prompt Discovery or Regression Avoidance?}
\label{subsection:validation-effectiveness}

Critic-based evaluation with correct training in Web of Lies showed one of the greatest performance improvements when using validated training over naive training, but many other experiments did not show a performance increase with validation.
We further investigate potential reasons for this.

Validated selection likely impacts performance using one of two mechanisms: either by preventing prompt $\theta$ from being replaced by a generated prompt $\theta'$ that decreases performance (\textbf{Regression Avoidance}), or by generating many prompts $\theta'_1, \ldots, \theta'_n$, with $n$ chances for a new prompt to result in high performance (\textbf{Prompt Discovery}).

\label{subsubsection:validation-case-study-wol}

\begin{figure}
    \centering 
    \captionsetup{type=figure}
    \includegraphics[width=0.9\linewidth]{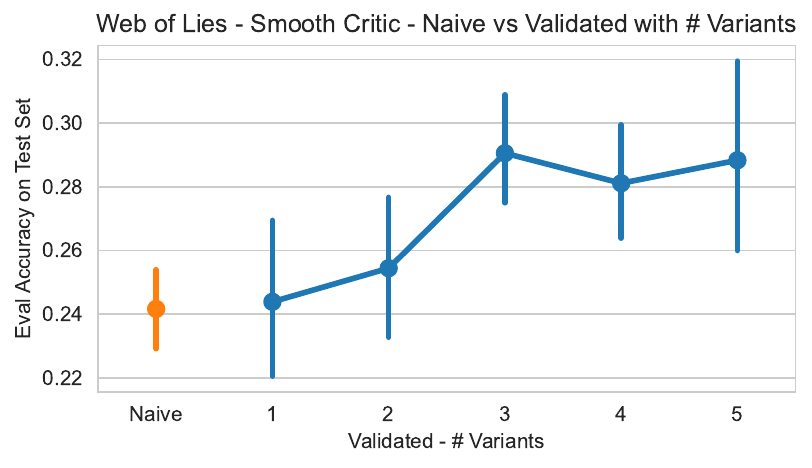}
    \caption{\label{figure:prompt-discovery}Performance in Web of Lies with naive selection, and validated selection using $n$ variants.}
\end{figure}

We instantiate 40 new trials of naive training in this domain, and 15 trials each of validated training with 1-5 generated variants, obtaining a power of $0.83$ for one-sided $t$-tests between naive and validated trials.
The results are shown in Figure \ref{figure:prompt-discovery}.

We find a significant improvement with validated prompt selection with 5 variants over naive selection ($p = 0.0014$), replicating our observations from Section \ref{subsection:variants-of-gradlike}.
However, we do not find a significant improvement from naive prompt selection to validated selection with one variant ($p=0.44$).

We test if the number of variants is correlated with increasing performance with validated selection, using a one-sided Kendall $\tau$ test.
We find that there is a significant, albeit somewhat weak, correlation ($\tau = 0.246$, $p=0.0025$).

Therefore, there is evidence to support that, at least in this configuration, the benefit of validated training comes primarily from prompt discovery rather than regression avoidance.

However, further research is required.
For example, in Section \ref{fig:overfit}, training for 100 batches should result in 100 opportunities for prompt discovery.
Given that performance is limited in those trials, regressions may be a factor in some configurations.

\section{Conclusion}

While the gradient metaphor has led to the development of successful APO methods, further investigation reveals that textual feedback generally does not behave like gradients.
Ablations that would decimate performance in a gradient-based system often have little impact on prompt optimization.
In some configurations, test accuracy improvements may stem from discovery of prevalence-hacking prompts.
The insights we present serve both to aid the selection of an appropriate APO method for a given task, and to motivate further understanding and development of future APO methods.

\section*{Limitations}

We note several limitations of our evaluation.
First, we restricted our investigation to prompt optimization, rather than in-context-example selection \citep{wan2024teachbettersmarterinstructions}.
We refer interested readers to \citet{min-etal-2022-rethinking} and \citet{pan-etal-2023-context} for insights about using examples with random labels.

Second, we note that there was high variance in our results, even with just one model and implementation of the APO framework.
While we attempt to mitigate this high variance by using many trials, the results may differ across multiple models, starting prompts, and meta-prompts used for prompt improvement.
Our power analyses usually assume an effect size of $d=0.8$, but the variants we test may have a smaller effect size, resulting in our tests not being able to observe a significant difference.
While the datasets include both tasks weighted towards internal knowledge (GPQA), and tasks weighted towards reasoning over the context (Web of Lies and Multistep Arithmetic), the results could be made more robust with additional evaluations, such as in coding or tool-use tasks.

Finally, LLMs have continued to become more capable with each iteration. 
A language model from 2021 may not have been able to follow the complex instructions required for gradient-like APO, and prompt generation methods such as backtranslation appear archaic just a few years later.

Some of the conclusions from this paper contradict those found in prior work using earlier models, such as \citet{yang2024largelanguagemodelsoptimizers}.
They find that performance with prompt optimization continues to improve for many optimization steps, rather than our observation in Figure \ref{fig:overfit} that training accuracy plateaus quickly.
While earlier language models may perform more incremental changes for each APO step, or may be more sensitive to exact prompt wording, modern LLMs synthesize a fairly complex prompt on the first step.

Based on this, the results in this paper may need to be re-evaluated for future generations of models.

\section*{Ethical Considerations}

Like many methods in this field, automatic prompt optimization may be used for either beneficial or harmful purposes.
Proper guardrails in the underlying language models provide a defense against using prompt optimization for nefarious goals.
However, these guardrails do not protect against the risk that a well-intentioned user may be overly confident in a model's output.

Foundation model usage also entails an environmental and resource usage impact.
Prompt optimization often uses fewer resources compared to model fine-tuning, but may still require a considerable number of tokens, especially in validated configurations.

\bibliography{bibliography}

\begin{thebibliography}{32}
\providecommand{\natexlab}[1]{#1}

\bibitem[{{Anthropic}(2025)}]{anthropic2025promptengineering}
{Anthropic}. 2025.
\newblock Prompt engineering overview.
\newblock Anthropic Documentation, Building with Claude series.
\newblock Available online at
  \url{https://docs.anthropic.com/en/docs/build-with-claude/prompt-engineering/overview}.

\bibitem[{Castelvecchi(2025)}]{Castelvecchi_2025}
Davide Castelvecchi. 2025.
\newblock \href {https://doi.org/10.1038/d41586-025-00406-7} {Deepmind ai
  crushes tough maths problems on par with top human solvers}.
\newblock \emph{Nature}, 638(8051):589–589.

\bibitem[{Cheng et~al.(2024)Cheng, Nie, and
  Swaminathan}]{cheng2024traceautodiffgenerativeoptimization}
Ching-An Cheng, Allen Nie, and Adith Swaminathan. 2024.
\newblock \href {https://arxiv.org/abs/2406.16218} {Trace is the next autodiff:
  Generative optimization with rich feedback, execution traces, and llms}.
\newblock \emph{Preprint}, arXiv:2406.16218.

\bibitem[{Do et~al.(2024)Do, Zhao, Brown, Xie, Zhao, Chen, Kawaguchi, Shieh,
  and He}]{do2024promptoptimizationadversarialincontext}
Xuan~Long Do, Yiran Zhao, Hannah Brown, Yuxi Xie, James~Xu Zhao, Nancy~F. Chen,
  Kenji Kawaguchi, Michael Shieh, and Junxian He. 2024.
\newblock \href {https://arxiv.org/abs/2312.02614} {Prompt optimization via
  adversarial in-context learning}.
\newblock \emph{Preprint}, arXiv:2312.02614.

\bibitem[{Fernando et~al.(2023)Fernando, Banarse, Michalewski, Osindero, and
  Rocktäschel}]{fernando2023promptbreederselfreferentialselfimprovementprompt}
Chrisantha Fernando, Dylan Banarse, Henryk Michalewski, Simon Osindero, and Tim
  Rocktäschel. 2023.
\newblock \href {https://arxiv.org/abs/2309.16797} {Promptbreeder:
  Self-referential self-improvement via prompt evolution}.
\newblock \emph{Preprint}, arXiv:2309.16797.

\bibitem[{Goodfellow et~al.(2014)Goodfellow, Pouget-Abadie, Mirza, Xu,
  Warde-Farley, Ozair, Courville, and
  Bengio}]{goodfellow2014generativeadversarialnetworks}
Ian~J. Goodfellow, Jean Pouget-Abadie, Mehdi Mirza, Bing Xu, David
  Warde-Farley, Sherjil Ozair, Aaron Courville, and Yoshua Bengio. 2014.
\newblock \href {https://arxiv.org/abs/1406.2661} {Generative adversarial
  networks}.
\newblock \emph{Preprint}, arXiv:1406.2661.

\bibitem[{Guo et~al.(2025)Guo, Wang, Guo, Li, Song, Tan, Liu, Bian, and
  Yang}]{guo2025evopromptconnectingllmsevolutionary}
Qingyan Guo, Rui Wang, Junliang Guo, Bei Li, Kaitao Song, Xu~Tan, Guoqing Liu,
  Jiang Bian, and Yujiu Yang. 2025.
\newblock \href {https://arxiv.org/abs/2309.08532} {Evoprompt: Connecting llms
  with evolutionary algorithms yields powerful prompt optimizers}.
\newblock \emph{Preprint}, arXiv:2309.08532.

\bibitem[{Gupta et~al.(2024)Gupta, Kirtania, Singha, Gulwani, Radhakrishna,
  Shi, and Soares}]{gupta2024metareflectionlearninginstructionslanguage}
Priyanshu Gupta, Shashank Kirtania, Ananya Singha, Sumit Gulwani, Arjun
  Radhakrishna, Sherry Shi, and Gustavo Soares. 2024.
\newblock \href {https://arxiv.org/abs/2405.13009} {Metareflection: Learning
  instructions for language agents using past reflections}.
\newblock \emph{Preprint}, arXiv:2405.13009.

\bibitem[{Jiang et~al.(2020)Jiang, Xu, Araki, and
  Neubig}]{jiang2020knowlanguagemodelsknow}
Zhengbao Jiang, Frank~F. Xu, Jun Araki, and Graham Neubig. 2020.
\newblock \href {https://arxiv.org/abs/1911.12543} {How can we know what
  language models know?}
\newblock \emph{Preprint}, arXiv:1911.12543.

\bibitem[{Kazemi et~al.(2025)Kazemi, Fatemi, Bansal, Palowitch, Anastasiou,
  Mehta, Jain, Aglietti, Jindal, Chen, Dikkala, Tyen, Liu, Shalit, Chiappa,
  Olszewska, Tay, Tran, Le, and Firat}]{kazemi2025bigbenchextrahard}
Mehran Kazemi, Bahare Fatemi, Hritik Bansal, John Palowitch, Chrysovalantis
  Anastasiou, Sanket~Vaibhav Mehta, Lalit~K. Jain, Virginia Aglietti, Disha
  Jindal, Peter Chen, Nishanth Dikkala, Gladys Tyen, Xin Liu, Uri Shalit,
  Silvia Chiappa, Kate Olszewska, Yi~Tay, Vinh~Q. Tran, Quoc~V. Le, and Orhan
  Firat. 2025.
\newblock \href {https://arxiv.org/abs/2502.19187} {Big-bench extra hard}.
\newblock \emph{Preprint}, arXiv:2502.19187.

\bibitem[{Khattab et~al.(2023)Khattab, Singhvi, Maheshwari, Zhang, Santhanam,
  Vardhamanan, Haq, Sharma, Joshi, Moazam, Miller, Zaharia, and
  Potts}]{khattab2023dspycompilingdeclarativelanguage}
Omar Khattab, Arnav Singhvi, Paridhi Maheshwari, Zhiyuan Zhang, Keshav
  Santhanam, Sri Vardhamanan, Saiful Haq, Ashutosh Sharma, Thomas~T. Joshi,
  Hanna Moazam, Heather Miller, Matei Zaharia, and Christopher Potts. 2023.
\newblock \href {https://arxiv.org/abs/2310.03714} {Dspy: Compiling declarative
  language model calls into self-improving pipelines}.
\newblock \emph{Preprint}, arXiv:2310.03714.

\bibitem[{Min et~al.(2022)Min, Lyu, Holtzman, Artetxe, Lewis, Hajishirzi, and
  Zettlemoyer}]{min-etal-2022-rethinking}
Sewon Min, Xinxi Lyu, Ari Holtzman, Mikel Artetxe, Mike Lewis, Hannaneh
  Hajishirzi, and Luke Zettlemoyer. 2022.
\newblock \href {https://doi.org/10.18653/v1/2022.emnlp-main.759} {Rethinking
  the role of demonstrations: What makes in-context learning work?}
\newblock In \emph{Proceedings of the 2022 Conference on Empirical Methods in
  Natural Language Processing}, pages 11048--11064, Abu Dhabi, United Arab
  Emirates. Association for Computational Linguistics.

\bibitem[{Mishra et~al.(2022)Mishra, Khashabi, Baral, Choi, and
  Hajishirzi}]{mishra2022reframinginstructionalpromptsgptks}
Swaroop Mishra, Daniel Khashabi, Chitta Baral, Yejin Choi, and Hannaneh
  Hajishirzi. 2022.
\newblock \href {https://arxiv.org/abs/2109.07830} {Reframing instructional
  prompts to gptk's language}.
\newblock \emph{Preprint}, arXiv:2109.07830.

\bibitem[{{OpenAI}(2025)}]{openai2025promptengineering}
{OpenAI}. 2025.
\newblock Prompt engineering guide.
\newblock OpenAI Platform Documentation, Guides.
\newblock Available online at
  \url{https://platform.openai.com/docs/guides/prompt‑engineering}.

\bibitem[{Pan et~al.(2023)Pan, Gao, Chen, and Chen}]{pan-etal-2023-context}
Jane Pan, Tianyu Gao, Howard Chen, and Danqi Chen. 2023.
\newblock \href {https://doi.org/10.18653/v1/2023.findings-acl.527} {What
  in-context learning ``learns'' in-context: Disentangling task recognition and
  task learning}.
\newblock In \emph{Findings of the Association for Computational Linguistics:
  ACL 2023}, pages 8298--8319, Toronto, Canada. Association for Computational
  Linguistics.

\bibitem[{Parthasarathy et~al.(2024)Parthasarathy, Zafar, Khan, and
  Shahid}]{parthasarathy2024ultimateguidefinetuningllms}
Venkatesh~Balavadhani Parthasarathy, Ahtsham Zafar, Aafaq Khan, and Arsalan
  Shahid. 2024.
\newblock \href {https://arxiv.org/abs/2408.13296} {The ultimate guide to
  fine-tuning llms from basics to breakthroughs: An exhaustive review of
  technologies, research, best practices, applied research challenges and
  opportunities}.
\newblock \emph{Preprint}, arXiv:2408.13296.

\bibitem[{Prasad et~al.(2023)Prasad, Hase, Zhou, and
  Bansal}]{prasad2023gripsgradientfreeeditbasedinstruction}
Archiki Prasad, Peter Hase, Xiang Zhou, and Mohit Bansal. 2023.
\newblock \href {https://arxiv.org/abs/2203.07281} {Grips: Gradient-free,
  edit-based instruction search for prompting large language models}.
\newblock \emph{Preprint}, arXiv:2203.07281.

\bibitem[{Pryzant et~al.(2023)Pryzant, Iter, Li, Lee, Zhu, and
  Zeng}]{pryzant2023automaticpromptoptimizationgradient}
Reid Pryzant, Dan Iter, Jerry Li, Yin~Tat Lee, Chenguang Zhu, and Michael Zeng.
  2023.
\newblock \href {https://arxiv.org/abs/2305.03495} {Automatic prompt
  optimization with "gradient descent" and beam search}.
\newblock \emph{Preprint}, arXiv:2305.03495.

\bibitem[{Ramnath et~al.(2025)Ramnath, Zhou, Guan, Mishra, Qi, Shen, Wang, Woo,
  Jeoung, Wang, Wang, Ding, Lu, Xu, Zhou, Srinivasan, Yan, Chen, Ding, Xu, and
  Cheong}]{ramnath2025systematicsurveyautomaticprompt}
Kiran Ramnath, Kang Zhou, Sheng Guan, Soumya~Smruti Mishra, Xuan Qi, Zhengyuan
  Shen, Shuai Wang, Sangmin Woo, Sullam Jeoung, Yawei Wang, Haozhu Wang, Han
  Ding, Yuzhe Lu, Zhichao Xu, Yun Zhou, Balasubramaniam Srinivasan, Qiaojing
  Yan, Yueyan Chen, Haibo Ding, and 2 others. 2025.
\newblock \href {https://arxiv.org/abs/2502.16923} {A systematic survey of
  automatic prompt optimization techniques}.
\newblock \emph{Preprint}, arXiv:2502.16923.

\bibitem[{Rein et~al.(2023)Rein, Hou, Stickland, Petty, Pang, Dirani, Michael,
  and Bowman}]{rein2023gpqagraduatelevelgoogleproofqa}
David Rein, Betty~Li Hou, Asa~Cooper Stickland, Jackson Petty, Richard~Yuanzhe
  Pang, Julien Dirani, Julian Michael, and Samuel~R. Bowman. 2023.
\newblock \href {https://arxiv.org/abs/2311.12022} {Gpqa: A graduate-level
  google-proof q\&a benchmark}.
\newblock \emph{Preprint}, arXiv:2311.12022.

\bibitem[{Robeyns et~al.(2025)Robeyns, Szummer, and
  Aitchison}]{robeyns2025selfimprovingcodingagent}
Maxime Robeyns, Martin Szummer, and Laurence Aitchison. 2025.
\newblock \href {https://arxiv.org/abs/2504.15228} {A self-improving coding
  agent}.
\newblock \emph{Preprint}, arXiv:2504.15228.

\bibitem[{Shin et~al.(2020)Shin, Razeghi, IV, Wallace, and
  Singh}]{shin2020autopromptelicitingknowledgelanguage}
Taylor Shin, Yasaman Razeghi, Robert L.~Logan IV, Eric Wallace, and Sameer
  Singh. 2020.
\newblock \href {https://arxiv.org/abs/2010.15980} {Autoprompt: Eliciting
  knowledge from language models with automatically generated prompts}.
\newblock \emph{Preprint}, arXiv:2010.15980.

\bibitem[{Vickerstaff et~al.(2019)Vickerstaff, Omar, and
  Ambler}]{Vickerstaff_Omar_Ambler_2019}
Victoria Vickerstaff, Rumana~Z. Omar, and Gareth Ambler. 2019.
\newblock \href {https://doi.org/10.1186/s12874-019-0754-4} {Methods to adjust
  for multiple comparisons in the analysis and sample size calculation of
  randomised controlled trials with multiple primary outcomes}.
\newblock \emph{BMC medical research methodology}, 19(1):129.

\bibitem[{Wan et~al.(2024)Wan, Sun, Nakhost, and
  Arik}]{wan2024teachbettersmarterinstructions}
Xingchen Wan, Ruoxi Sun, Hootan Nakhost, and Sercan~O. Arik. 2024.
\newblock \href {https://arxiv.org/abs/2406.15708} {Teach better or show
  smarter? on instructions and exemplars in automatic prompt optimization}.
\newblock \emph{Preprint}, arXiv:2406.15708.

\bibitem[{Wang et~al.(2025)Wang, Li, Song, Xu, Tang, Zhuge, Pan, Song, Li,
  Singh, Tran, Li, Ma, Zheng, Qian, Shao, Muennighoff, Zhang, Hui, Lin,
  Brennan, Peng, Ji, and Neubig}]{wang2025openhandsopenplatformai}
Xingyao Wang, Boxuan Li, Yufan Song, Frank~F. Xu, Xiangru Tang, Mingchen Zhuge,
  Jiayi Pan, Yueqi Song, Bowen Li, Jaskirat Singh, Hoang~H. Tran, Fuqiang Li,
  Ren Ma, Mingzhang Zheng, Bill Qian, Yanjun Shao, Niklas Muennighoff, Yizhe
  Zhang, Binyuan Hui, and 5 others. 2025.
\newblock \href {https://arxiv.org/abs/2407.16741} {Openhands: An open platform
  for ai software developers as generalist agents}.
\newblock \emph{Preprint}, arXiv:2407.16741.

\bibitem[{Xu et~al.(2022)Xu, Chen, Du, Shao, Wang, Li, and
  Yang}]{xu2022gpsgeneticpromptsearch}
Hanwei Xu, Yujun Chen, Yulun Du, Nan Shao, Yanggang Wang, Haiyu Li, and Zhilin
  Yang. 2022.
\newblock \href {https://arxiv.org/abs/2210.17041} {Gps: Genetic prompt search
  for efficient few-shot learning}.
\newblock \emph{Preprint}, arXiv:2210.17041.

\bibitem[{Yang et~al.(2024)Yang, Wang, Lu, Liu, Le, Zhou, and
  Chen}]{yang2024largelanguagemodelsoptimizers}
Chengrun Yang, Xuezhi Wang, Yifeng Lu, Hanxiao Liu, Quoc~V. Le, Denny Zhou, and
  Xinyun Chen. 2024.
\newblock \href {https://arxiv.org/abs/2309.03409} {Large language models as
  optimizers}.
\newblock \emph{Preprint}, arXiv:2309.03409.

\bibitem[{Yin et~al.(2025)Yin, Wang, Pan, Lin, Wan, and
  Wang}]{yin2025godelagentselfreferentialagent}
Xunjian Yin, Xinyi Wang, Liangming Pan, Li~Lin, Xiaojun Wan, and William~Yang
  Wang. 2025.
\newblock \href {https://arxiv.org/abs/2410.04444} {G\"odel agent: A
  self-referential agent framework for recursive self-improvement}.
\newblock \emph{Preprint}, arXiv:2410.04444.

\bibitem[{Yuksekgonul et~al.(2025)Yuksekgonul, Bianchi, Boen, Liu, Lu, Huang,
  Guestrin, and Zou}]{yuksekgonul2024textgrad}
Mert Yuksekgonul, Federico Bianchi, Joseph Boen, Sheng Liu, Pan Lu, Zhi Huang,
  Carlos Guestrin, and James Zou. 2025.
\newblock Optimizing generative ai by backpropagating language model feedback.
\newblock \emph{Nature}, 639:609--616.

\bibitem[{Zelikman et~al.(2024)Zelikman, Lorch, Mackey, and
  Kalai}]{zelikman2024selftaughtoptimizerstoprecursively}
Eric Zelikman, Eliana Lorch, Lester Mackey, and Adam~Tauman Kalai. 2024.
\newblock \href {https://arxiv.org/abs/2310.02304} {Self-taught optimizer
  (stop): Recursively self-improving code generation}.
\newblock \emph{Preprint}, arXiv:2310.02304.

\bibitem[{Zhang et~al.(2025)Zhang, Hu, Lu, Lange, and
  Clune}]{zhang2025darwingodelmachineopenended}
Jenny Zhang, Shengran Hu, Cong Lu, Robert Lange, and Jeff Clune. 2025.
\newblock \href {https://arxiv.org/abs/2505.22954} {Darwin godel machine:
  Open-ended evolution of self-improving agents}.
\newblock \emph{Preprint}, arXiv:2505.22954.

\bibitem[{Zhou et~al.(2023)Zhou, Muresanu, Han, Paster, Pitis, Chan, and
  Ba}]{zhou2023largelanguagemodelshumanlevel}
Yongchao Zhou, Andrei~Ioan Muresanu, Ziwen Han, Keiran Paster, Silviu Pitis,
  Harris Chan, and Jimmy Ba. 2023.
\newblock \href {https://arxiv.org/abs/2211.01910} {Large language models are
  human-level prompt engineers}.
\newblock \emph{Preprint}, arXiv:2211.01910.

\end{thebibliography}

\appendix

\section{Resource Usage}
\label{appendix:token-usage}

Our token counts are broken down as follows:

\textbf{Section \ref{subsection:textual-gradient-vs-mc-onestep}} 619838771 input tokens for combined experiments; 638054972 output tokens.

\textbf{Section \ref{subsection:variants-of-gradlike}}
1648101679 input tokens; 877691530 output tokens.

\textbf{Section \ref{subsection:overfit}}
155436119 input tokens and 44979764 output tokens for training; 316769262 input and 238513122 output tokens for evaluation.

\textbf{Section \ref{subsection:bbehwol-case-study}} 16071000 input tokens; 18614182 output tokens (for the prompt ablation study).

\textbf{Section \ref{subsubsection:validation-case-study-wol}} 826090922 input tokens; 301464758 output tokens.

\textbf{Total} 3582307753 input tokens; 2119318328 output tokens.

Due to a shared account, it is difficult to attribute exact costs. 
At on-demand token rates at time of writing (\$3 per million input tokens; \$15 per million output tokens), the experiments cost \$10746.92 in input tokens and \$31789.77 in output tokens, for a total of \$42536.69.

We estimate that a similar number of tokens were used in preliminary experiments; our original research idea relied on strong assumptions about the gradient hypothesis, and we attempted numerous variations of those experiments before investigating whether the gradient hypothesis itself was true.

There was an additional, minor, cost for an AWS EC2 instance (non-GPU) to host the program containing the API client, and an EFS filesystem to store experiment results, but these costs were not closely tracked.

\section{Extended Results}

\label{appendix:p-values}

We include $p$-values for all comparisons in this paper.
Note that due to our use of non-parametric permutation tests (using 100000 permutations), the $p$-values vary slightly each time they are calculated.

\textbf{Section 3.1 RQ1}

Two-sided permutation t-test; comparing gradient-like vs one-step feedback driven APO; correct evaluations.

{\centering
\begin{tabular}{l|ll}
 & Naive & Validated \\\hline
GPQA & 0.32596 & 0.45238 \\
MSA & 0.02918 & 0.96825 \\
WoL & 0.00350 & 0.76190 \\
\end{tabular}\par}

\textbf{Section 3.1 RQ2}

One-sided permutation t-test; comparing correct evaluations with incorrect evaluations.

{\centering
\begin{tabular}{ll|ll}
 &  & Gradient-like & One-step \\\hline
GPQA & Naive & 0.25413 & 0.86616 \\
GPQA & Validated & 0.73413 & 0.22222 \\
MSA & Naive & 0.23197 & 0.39818 \\
MSA & Validated & 0.93651 & 0.19841 \\
WoL & Naive & 0.25818 & 0.67516 \\
WoL & Validated & 0.77778 & 0.45238 \\
\end{tabular}\par}

One-sided permutation t-test; comparing correct evaluations with missing evaluations.

{\centering
\begin{tabular}{ll|ll}
 &  & Gradient-like & One-step \\\hline
GPQA & Naive & 0.94817 & 0.87397 \\
GPQA & Validated & 0.74603 & 0.65476 \\
MSA & Naive & 0.05621 & 0.94999 \\
MSA & Validated & 0.38492 & 0.63492 \\
WoL & Naive & 0.00006 & 0.00191 \\
WoL & Validated & 0.61905 & 0.46032 \\
\end{tabular}\par}

\textbf{Section 3.1 RQ3}

One-sided permutation t-test; naive vs. validated.

{\centering
\begin{tabular}{l|l}
\textbf{GPQA} & \textbf{$p$-value} \\\hline
Prompt-only (rewrite) & 0.83252 \\
Prompt-only (improve) & 0.52184 \\

Gradient-like & 0.92019 \\
Gradient-like (incor. eval) & 0.59086 \\
Gradient-like (no eval) & 0.74801 \\
One-step & 0.95823 \\
One-step (incor. eval) & 0.99987 \\
One-step (no eval) & 0.93864 \\
\end{tabular}\par}

{\centering
\begin{tabular}{l|l}
\textbf{Multistep Arithmetic} & \textbf{$p$-value} \\\hline
Prompt-only (rewrite) & 0.28006 \\
Prompt-only (improve) & 0.15365 \\
Gradient-like & 0.23385 \\
Gradient-like (incor. eval) & 0.02993 \\
Gradient-like (no eval) & 0.04177 \\
One-step & 0.67234 \\
One-step (incor. eval) & 0.91584 \\
One-step (no eval) & 0.84075 \\
\end{tabular}\par}

{\centering
\begin{tabular}{l|l}
\textbf{Web of Lies} & \textbf{$p$-value} \\\hline
Prompt-only (rewrite) & 0.36356 \\
Prompt-only (improve) & 0.06098 \\
Gradient-like & 0.98010 \\
Gradient-like (incor. eval) & 0.93731 \\
Gradient-like (no eval) & 0.26386 \\
One-step & 0.48244 \\
One-step (incor. eval) & 0.65189 \\
One-step (no eval) & 0.03107 \\
\end{tabular}\par}

One-sided permutation t-test, validated with incorrect training and correct validation vs validated with incorrect training and incorrect validation.

{\centering
\begin{tabular}{l|ll}
 & Gradient-like & One-step \\\hline
GPQA & 0.15079 & 0.77778 \\
MSA & 0.02381 & 0.26984 \\
WoL & 0.50000 & 0.54762 \\
\end{tabular}\par}

\textbf{Section 3.1 RQ4}

One-sided permutation t-test, Prompt-only generation vs. default prompt.

{\centering
\begin{tabular}{ll|ll}
 &  & Improve & Rewrite \\\hline
GPQA & Naive & 0.00001 & 0.43243 \\
GPQA & Validated & 0.00078 & 0.88646 \\
MSA & Naive & 0.00214 & 0.87150 \\
MSA & Validated & 0.00038 & 0.38130 \\
WoL & Naive & 0.99952 & 0.93143 \\
WoL & Validated & 0.40777 & 0.62740 \\
\end{tabular}\par}

One-sided permutation t-test, gradient-like with correct training vs. prompt-only improve.

{\centering
\begin{tabular}{l|ll}
 & Naive & Validated \\\hline
GPQA & 0.00451 & 0.46429 \\
MSA & 0.00001 & 0.00397 \\
WoL & 0.00001 & 0.56746 \\
\end{tabular}\par}

\textbf{Section 3.2}

One-sided permutation t-test; correct vs incorrect training data.

{\centering
\begin{tabular}{ll|ll}
 &  & Bin. Critic & Sm. Critic \\\hline
GPQA & Naive & 0.32143 & 0.02560 \\
GPQA & Valid. & 0.36905 & 0.83730 \\
MSA & Naive & 0.77870 & 0.54491 \\
MSA & Valid. & 0.77381 & 0.94841 \\
WoL & Naive & 0.43937 & 0.66560 \\
WoL & Valid. & 0.98413 & 0.08333 \\
\end{tabular}\par}

Two-sided permutation t-test; direct vs critic-based evaluation with correct training data.

{\centering
\begin{tabular}{ll|ll}
 &  & Bin. Critic & Sm. Critic \\\hline
GPQA & Naive & 0.96321 & 0.38652 \\
GPQA & Valid. & 0.27778 & 0.51587 \\
MSA & Naive & 0.88347 & 0.68141 \\
MSA & Valid. & 0.15079 & 0.08730 \\
WoL & Naive & 0.32646 & 0.00112 \\
WoL & Valid. & 0.42857 & 0.00794 \\
\end{tabular}\par}

Two-sided permutation t-test; direct vs critic-based evaluation with incorrect training data.

{\centering
\begin{tabular}{ll|ll}
 &  & Bin. Critic & Sm. Critic \\\hline
GPQA & Naive & 0.75623 & 0.69617 \\
GPQA & Valid. & 0.60317 & 0.23016 \\
MSA & Naive & 0.21124 & 0.29172 \\
MSA & Valid. & 0.15873 & 0.50000 \\
WoL & Naive & 0.66823 & 0.05724 \\
WoL & Valid. & 0.01587 & 0.01587 \\
\end{tabular}\par}

One-sided permutation t-test; naive vs validated training.

{\centering
\begin{tabular}{ll|ll}
 & Training & Bin. Critic & Sm. Critic \\\hline
GPQA & Correct & 0.38697 & 0.84701 \\
GPQA & Incorrect & 0.53860 & 0.05930 \\
MSA & Correct & 0.40689 & 0.54588 \\
MSA & Incorrect & 0.39541 & 0.09246 \\
WoL & Correct & 0.76608 & 0.00002 \\
WoL & Incorrect & 0.02292 & 0.00080 \\
\end{tabular}\par}

One-sided permutation t-test; validated training with incorrect training data; correct vs incorrect validation data.

{\centering
\begin{tabular}{l|ll}
 & Bin. Critic & Sm. Critic \\\hline
GPQA & 0.61111 & 0.78175 \\
MSA & 0.92857 & 0.08730 \\
WoL & 0.05556 & 0.07937 \\
\end{tabular}\par}

\section{Case Study}

\label{appendix:case-study-text}

We include expanded versions of all steps of the case study.

\subsection{Generator Prompt}

The following prompt performed well on Web of Lies, and was thus used in the case study in Section \ref{subsection:bbehwol-case-study}:

\begin{displayquote}
    \small
    \obeylines
    When answering logic puzzles or complex questions:

1. Begin by systematically organizing all relevant information:
   - For puzzles involving people, create a comprehensive tracking table showing each person and their current status (Truth/Lie/Unknown)
   - For truth-teller/liar puzzles:
     * Immediately identify and clearly mark all contradiction pairs (where two people make opposing claims) - exactly one tells truth
     * Immediately identify and clearly mark all mutual endorsement pairs (where two people validate each other) - either both tell truth or both lie
     * Update your tracking table after EACH new deduction
     * For complex puzzles with many individuals, create a relationship graph showing how statements interconnect
   - Explicitly label any given facts in the problem as "anchor points" to build from
   - Pay special attention to the specific individuals mentioned in the question - these MUST be determined

2. Develop a systematic and exhaustive approach for truth-teller/liar puzzles:
   - Start with established "anchor points" (known truth-tellers/liars) and methodically propagate implications
   - For each contradiction pair (A says B lies, B says A lies):
     * Explicitly state that exactly one tells truth
     * Test both possibilities if necessary
   - For each mutual endorsement pair (A says B tells truth, B says A tells truth):
     * Explicitly state that either both tell truth or both lie
     * Test both possibilities if necessary
   - For complex statements like "exactly two of X, Y, Z tell the truth":
     * Create mini truth tables showing all possible T/F combinations
     * Eliminate combinations that violate known constraints
     * Explicitly show which combinations remain valid
   - After EACH deduction, immediately:
     * Update your tracking table
     * Check if the new deduction creates implications for other people
     * Verify the deduction is consistent with all previously established facts
   - For complex interconnected statements, create a dependency chain showing how each deduction affects others
   - When you encounter a contradiction, explicitly state:
     * Which assumption led to the contradiction
     * Why it's a contradiction
     * What alternative assumption you'll now explore

3. Be EXTREMELY persistent and thorough - NEVER accept "unknown" prematurely:
   - CRITICAL: These puzzles are ALWAYS designed to have definitive answers - if you're tempted to conclude "unknown," this is a STRONG SIGNAL that you need to dig deeper
   - When you think you've reached a dead end, systematically review all statements again
   - When facing difficult deductions, try ALL of these approaches before concluding "unknown":
     * Assumption testing: Temporarily assume someone tells truth/lies and follow the logical consequences
     * If an assumption leads to a contradiction, explicitly state why the opposite must be true
     * For EVERY individual that seems undetermined, test BOTH truth/lie assumptions
     * For complex group statements, try ALL possible truth value combinations systematically
   - When facing multiple possibilities, explore EACH branch separately and thoroughly
   - Use process of elimination rigorously - if all but one possibility leads to contradiction, the remaining one must be true
   - For apparent "unknowns," explicitly check if combining information from multiple statements creates new constraints
   - If direct deduction isn't working, look for longer chains of implications
   - NEVER give up on determining someone's status - if you're stuck, this means you need to explore more chains of reasoning

4. Show your complete reasoning process with no gaps:
   - Document EACH step in your chain of logical deductions
   - Explain WHY each conclusion follows from previous deductions
   - NEVER state a conclusion without showing the FULL logical path that led there
   - For complex deductions, break down the reasoning into smaller, more manageable steps
   - Periodically summarize what has been established so far
   - Use clear language to indicate logical relationships (if/then, therefore, because, etc.)
   - Make logical dependencies explicit through clear formatting

5. Implement rigorous verification and cross-checking:
   - After each key deduction, verify it by testing if the opposite assumption leads to a contradiction
   - Regularly verify that your current set of truth/lie assignments is consistent with ALL statements
   - For each person whose status you determine, explicitly check this against ALL statements involving them
   - Before finalizing your answer, verify that your solution is consistent with EVERY statement in the puzzle
   - Pay special attention to verifying the status of individuals mentioned in the question
   - If you find a contradiction in your solution, explicitly backtrack to identify where the reasoning went wrong
   - Create a final verification section showing how your conclusions are consistent with all statements

6. Present your answer clearly:
   - Include a clear summary section that recaps the key deductions for each person mentioned in the question
   - Show exactly how you determined the status of each person mentioned in the question
   - Format your final answer exactly as requested in the problem
   - Double-check that your final set of truth/lie assignments is consistent with all statements in the problem
   - For the specific individuals mentioned in the question, provide explicit verification of their status

7. Special techniques for breaking through apparent dead ends:
   - If direct deduction isn't working, try assumption testing more aggressively
   - When you feel tempted to label something as "unknown," this is a warning sign - redouble your efforts
   - Look for hidden constraints by combining multiple statements
   - For each "unknown" individual, explicitly try BOTH truth and lie assumptions and follow chains of implications
   - Check if determining one person's status would force another's through chains of statements
   - Focus on contradiction pairs and mutual endorsement pairs to break symmetries
   - For complex interconnected statements, create a dependency graph showing how each person's status affects others
   - Look for statements that severely constrain possibilities (like "exactly one person tells truth")
   - Remember that the individuals mentioned in the question ALWAYS have determinable truth values

Remember: NEVER declare a value "unknown" unless you have exhaustively explored ALL logical approaches and combinations. These puzzles are ALWAYS designed to have definitive answers, so if you're tempted to say "unknown," that's a definite sign you need to dig deeper and try more approaches. For the specific individuals mentioned in the question, you MUST determine their status - verify your conclusions about them multiple times before finalizing your answer.
\end{displayquote}

We traced this prompt back two iterations to the earliest prompt that contained a block substantially similar to block 3:

\begin{displayquote}
    \small
    \obeylines
    When answering logic puzzles or complex questions:

1. Begin by carefully organizing all relevant information:
   - For puzzles involving people, create a clear list of all individuals and their relevant attributes
   - For truth-teller/liar puzzles:
     * Create a tracking table showing each person's current status (Truth/Lie/Unknown)
     * Identify and mark all contradiction pairs (where two people make opposing claims)
     * Identify and mark all mutual endorsement pairs (where two people validate each other)
     * For complex puzzles with many individuals, create a relationship graph or matrix
   - Identify any explicit facts given in the problem as your "anchor points"
   - Pay special attention to the specific individuals mentioned in the question

2. Develop a systematic approach tailored to truth-teller/liar puzzles:
   - Start with established "anchor points" (known truth-tellers/liars) and propagate implications
   - For contradiction pairs (A says B lies, B says A lies), note that exactly one tells truth
   - For mutual endorsement pairs (A says B tells truth, B says A tells truth), note that either both tell truth or both lie
   - For complex statements like "exactly two of X, Y, Z tell the truth":
     * Create mini truth tables showing all possible T/F combinations
     * Eliminate combinations that violate known constraints
   - Regularly update your tracking table as new deductions are made
   - After each deduction, immediately check if it creates new implications for other people
   - For very complex puzzles, create periodic "checkpoints" where you summarize established facts

3. Be extremely persistent and thorough:
   - IMPORTANT: These puzzles are designed to have definitive answers - avoid concluding "unknown" prematurely
   - When you think you've reached a dead end, systematically review all statements again
   - If direct deduction isn't working, try assumption testing:
     * Temporarily assume someone tells truth/lies and follow the logical consequences
     * If an assumption leads to a contradiction, the opposite must be true
   - For complex group statements, try all possible truth value combinations systematically
   - When facing multiple possibilities, explore each branch separately to see if one leads to contradictions
   - Use process of elimination rigorously - if all but one possibility leads to contradiction, the remaining one must be true
   - Never give up until you've exhausted ALL possible logical approaches

4. Show your complete reasoning process:
   - Document each step in your chain of logical deductions
   - Explain why each conclusion follows from previous deductions or given information
   - Never jump to conclusions without showing the logical path that led there
   - Use indentation or bullet hierarchies to make logical dependencies clearer
   - Highlight critical turning points in your reasoning process

5. Implement verification and cross-checking:
   - Regularly verify that your current set of truth/lie assignments is consistent with all statements
   - When you reach a conclusion about a person, double-check by testing it against all relevant statements
   - For key deductions, try the opposite assumption to confirm it leads to a contradiction
   - Before finalizing your answer, verify that your solution is consistent with EVERY statement in the puzzle
   - If you find a contradiction in your solution, backtrack to identify where the reasoning went wrong

6. Present your answer clearly:
   - Include a clear summary section that recaps the key deductions that led to your final conclusion
   - Format your final answer exactly as requested in the problem
   - Verify that your answer matches the deductions you made
   - For the specific individuals mentioned in the question, ensure you have definitive answers (not "unknown")
   - Double-check that your final set of truth/lie assignments is consistent with all statements in the problem

7. Special techniques for managing complex puzzles:
   - Use visualization techniques like truth tables, matrices, or graphs to track relationships
   - Break the puzzle into connected components and solve each component separately
   - Look for "forcing moves" - deductions that must be true regardless of other unknowns
   - Track both what is known and what is ruled out
   - When the puzzle seems too complex, focus on the specific individuals mentioned in the question
   - Look for chains of implications that connect known facts to the individuals in question

Remember: NEVER declare a value "unknown" unless you have exhaustively explored all logical approaches and combinations. These puzzles are designed to have definitive answers, so if you're tempted to say "unknown," that's a sign you need to dig deeper.
\end{displayquote}

\subsection{Evaluator output and feedback}

The following is the evaluator output ($\evaluationShort$) for one of the questions, in the iteration that led to the production of the previous prompt:

\begin{displayquote}
    \small
    \obeylines
    I'm evaluating a solution to a complex truth-teller/liar puzzle where the goal is to determine whether Raymond, Murphy, and Sima tell the truth.

The solution correctly identifies some contradictory pairs and establishes that Tamika is a truth-teller as directly stated in the problem. It then makes some progress by determining that Oda and Oprah must be truth-tellers based on their statements and Alexis's statement.

However, the solution fails to reach definitive conclusions about Raymond, Murphy, and Sima, stating that there are "too many variables and not enough constraints" to determine their truth values. This is incorrect - the puzzle does provide sufficient information to determine the truth values of all three people.

The solution stops short of making crucial deductions:

1. For Raymond: The solution identifies Dallas's statement about Raymond but doesn't follow through with determining Raymond's status.

2. For Murphy: The solution notes that Murphy and Melody make statements about each other but doesn't resolve whether they're truth-tellers or liars.

3. For Sima: The solution mentions Sima's statement but doesn't use it effectively in a chain of reasoning.

The correct answer is that Raymond, Murphy, and Sima all tell the truth (yes, yes, yes). This can be determined through careful analysis of the statements and their implications, particularly by working with the known truth-tellers (Tamika, Oda, Oprah) and following the logical chains from there.

Since the final answer provided is incorrect (unknown, unknown, unknown), the score must be 0, regardless of the partial progress made in the reasoning.

$<$SCORE$>$0$<$/SCORE$>$
\end{displayquote}

This led to the following feedback about the output ($\feedbackForOutput$):

\begin{displayquote}
    \small
    \obeylines
    After analyzing the three examples, I can see a consistent pattern in how the solutions are being evaluated. All three solutions received a score of 0/100 for the same fundamental reason: they concluded that the truth values of the target individuals were "unknown" when in fact they were determinable from the given information.

Key issues with all three solutions:

1. **Insufficient persistence in logical deduction**: In each case, the solver correctly identified some initial relationships (contradiction pairs, mutual endorsement pairs) and made some valid deductions, but stopped short of pursuing all logical chains to their conclusion. The evaluator consistently notes that with more thorough analysis, definitive answers could have been reached.

2. **Premature declaration of "unknown"**: All three solutions gave up too easily and declared the truth values as "unknown" when further logical steps would have yielded definitive answers. The puzzles were designed to have unique solutions.

3. **Failure to fully leverage known facts**: While the solutions correctly identified anchor points (like "Dallas lies" or "Sal tells the truth"), they didn't fully exploit these to create chains of deductions that would eventually determine the truth values of the target individuals.

4. **Incomplete exploration of logical implications**: The solutions identified many important relationships but didn't thoroughly explore what these relationships imply when combined with other statements in the puzzle.

To improve future outputs:

1. Be more persistent and thorough in following logical chains - don't give up until all possible deductions have been exhausted.

2. Create more systematic tracking of truth values as they're determined, and regularly check if newly determined values can help resolve previously undetermined ones.

3. When faced with multiple possibilities, try working with each possibility separately to see if one leads to a contradiction, which would eliminate it.

4. Only declare a truth value as "unknown" after exhaustively exploring all possible logical approaches and confirming there's genuinely insufficient information.

5. For complex puzzles, consider using more structured approaches like truth tables or matrices to track relationships between statements.

Remember that these logic puzzles are designed to have definitive answers, so if the solution seems incomplete, that's a sign to dig deeper rather than concluding the answer is unknowable.

\end{displayquote}

This was used to construct the following feedback about the prompt ($\feedbackForPrompt$):

\begin{displayquote}
    \small
    \obeylines
    Based on the examples and feedback, I can see that the current prompt needs significant improvement to better guide the system in solving complex truth-teller/liar puzzles. Here are my recommendations:

1. **Emphasize persistence and thoroughness**: The prompt should strongly emphasize that these puzzles are designed to have definitive answers, and the system should never give up and declare "unknown" until truly exhausting all logical possibilities.

2. **Add specific techniques for truth-teller/liar puzzles**: While the current prompt mentions some general approaches, it should include more specific techniques for these particular puzzles:
   - Creating truth tables for tracking the status of each person
   - Systematically propagating known truth values through chains of statements
   - Working with mutual endorsement and contradiction pairs as starting points
   - Techniques for resolving complex statements about groups (e.g., "exactly two of X, Y, Z tell the truth")

3. **Include verification steps**: Add explicit instructions to verify potential solutions by checking consistency with all statements in the puzzle.

4. **Add structured approach for handling uncertainty**: Include guidance on how to handle situations with multiple possibilities - such as exploring each branch separately to see if one leads to contradictions.

5. **Emphasize incremental progress tracking**: The prompt should stress the importance of regularly reviewing what has been established and checking if newly determined truth values can help resolve previously undetermined ones.

6. **Add explicit warning against premature "unknown" conclusions**: The prompt should explicitly warn against concluding that something is "unknown" before exhausting all logical approaches.

7. **Suggest visualization techniques**: Recommend creating visual representations (like graphs or matrices) to track complex relationships between statements.

8. **Add instructions for managing cognitive load**: Provide more specific guidance on breaking down complex puzzles into manageable parts while maintaining overall consistency.

These changes would help the system be more persistent, systematic, and thorough in solving these complex logic puzzles, avoiding the premature "unknown" conclusions seen in the examples.
\end{displayquote}

\section{Prompts used for Prompt Optimization}

\label{appendix:prompts}

\subsection{Prompt-Only}

The following prompt is used for Prompt-Only (rewrite) trials:

\begin{displayquote}
    \small
    \obeylines
    \input{prompts/prompt_only/rewrite.md}
\end{displayquote}

The following prompt is used for Prompt-Only (improve) trials:

\begin{displayquote}
    \small
    \obeylines
    \input{prompts/prompt_only/improve.md}
\end{displayquote}

\subsection{Prompt + Output + Evaluation}

These are the main prompts used in the APO loop.

\subsubsection{Gradlike}

Given the evaluation $\evaluationShort$, the following prompt produces output feedback $\feedbackForOutput$:

\begin{displayquote}
    \small
    \obeylines
    You are a system for improving example outputs based on feedback, part of a larger optimization system.
You will be provided with a set of examples as $<$INPUT$>$ $<$OUTPUT$>$ pairs.
Each pair was evaluated by an evaluator, with the task given in the $<$EVALUATOR-TASK$>$ tag.
The evaluator gives feedback in the $<$FEEDBACK$>$ tag.
Your task is to provide guidance on how to improve the output based on the feedback.
Do not update the prompt, and do not speculate on how to improve the prompt. Only discuss how to improve the example outputs.
After careful thought, provide feedback for each example output in $<$OUTPUT-FEEDBACK$>$ tags.
\end{displayquote}

The next prompt produces prompt feedback $\feedbackForPrompt$:

\begin{displayquote}
    \small
    \obeylines
    You are a system for suggesting improvements to a prompt (set of instructions) based on feedback, part of a larger optimization system.
You will be provided with a set of examples as $<$INPUT$>$ $<$OUTPUT$>$ pairs.
These pairs were generated by a task solver, with the instructions from the $<$PROMPT$>$ tag.
The task solver's task is given in the $<$TASK$>$ tag.
Suggestions to improve each output are provided in the $<$OUTPUT-FEEDBACK$>$ tag.
Your task is to provide guidance on how to improve the prompt based on the output feedback.
Do not update the prompt; only provide feedback about how to improve the prompt based on the $<$OUTPUT-FEEDBACK$>$.
After careful thought, provide feedback for the prompt in $<$PROMPT-FEEDBACK$>$ tags.
\end{displayquote}

Finally, the last prompt creates an updated prompt $P$:

\begin{displayquote}
    \small
    \obeylines
    You are a system for improving a prompt (set of instructions) based on feedback, part of a larger optimization system.
The current prompt is provided in $<$PROMPT$>$, and it is meant to solve the task given in the $<$TASK$>$ tag.
Feedback about how to improve the prompt is in the $<$PROMPT-FEEDBACK$>$ tag. 
Apply the feedback to the prompt to improve it.
After careful thought, output a single new prompt surrounded by $<$UPDATED-PROMPT$>$ tags.
\end{displayquote}

\subsubsection{One-Step}

Given the evaluation $\evaluationShort$, the following prompt produces a new prompt $\prompt$:

\begin{displayquote}
    \small
    \obeylines
    You are a system for improving the prompt (set of instructions) based on feedback from examples.
You will be provided the original prompt for a problem solver in $<$PROMPT$>$ tags.
The prompt is intended to help solve the task specified in the $<$TASK$>$ tags.
The result will be evaluated by an evaluator with the task given by $<$EVALUATOR-TASK$>$.
You must assist with improving the $<$PROMPT$>$ to solve the problem given by $<$TASK$>$.
A series of inputs, outputs, and evaluations are provided in $<$INPUT$>$, $<$OUTPUT$>$, and $<$EVALUATION$>$.
Feedback about each evaluation is provided in the $<$FEEDBACK$>$ tag.
After thinking about how to improve the prompt in response to the feedback, output a single new prompt surrounded by $<$UPDATED-PROMPT$>$ tags.
\end{displayquote}

\subsection{Prompt + Output}

These prompts are used when we don't include an evaluation, and ask the LLM to make up feedback to improve the generator anyways.

\subsubsection{Gradlike}

We first ask the LLM to invent feedback $\feedbackForOutput$, while only having access to $\promptOutput$ and not $\evaluationShort$:

\begin{displayquote}
    \small
    \obeylines
    You are a system for improving example outputs, part of a larger optimization system.
You will be provided with a set of examples as $<$INPUT$>$ $<$OUTPUT$>$ pairs.
Your task is to provide guidance on how to improve the output's quality and accuracy.
Do not update the prompt, and do not speculate on how to improve the prompt. Only discuss how to improve the example outputs.
After careful thought, provide feedback for each example output in $<$OUTPUT-FEEDBACK$>$ tags.
\end{displayquote}

The next prompt produces prompt feedback $\feedbackForPrompt$, using the hallucinated feedback from the previous step:

\begin{displayquote}
    \small
    \obeylines
    You are a system for suggesting improvements to a prompt (set of instructions) based on feedback, part of a larger optimization system.
You will be provided with a set of examples as $<$INPUT$>$ $<$OUTPUT$>$ pairs.
These pairs were generated by a task solver, with the instructions from the $<$PROMPT$>$ tag.
The task solver's task is given in the $<$TASK$>$ tag.
Suggestions to improve each output are provided in the $<$OUTPUT-FEEDBACK$>$ tag.
Your task is to provide guidance on how to improve the prompt based on the output feedback.
Do not update the prompt; only provide feedback about how to improve the prompt based on the $<$OUTPUT-FEEDBACK$>$.
After careful thought, provide feedback for the prompt in $<$PROMPT-FEEDBACK$>$ tags.
\end{displayquote}

Finally, the last prompt creates an updated prompt $\prompt'$:

\begin{displayquote}
    \small
    \obeylines
    You are a system for improving a prompt (set of instructions) based on feedback, part of a larger optimization system.
The current prompt is provided in $<$PROMPT$>$, and it is meant to solve the task given in the $<$TASK$>$ tag.
Feedback about how to improve the prompt is in the $<$PROMPT-FEEDBACK$>$ tag. 
Apply the feedback to the prompt to improve it.
After careful thought, output a single new prompt surrounded by $<$UPDATED-PROMPT$>$ tags.
\end{displayquote}

\subsubsection{One-Step}

This prompt asks the LLM to improve $\prompt$ directly to obtain $\prompt'$.

\begin{displayquote}
    \small
    \obeylines
    You are a system for improving the prompt (set of instructions) based on examples.
You will be provided the original prompt for a problem solver in $<$PROMPT$>$ tags.
The prompt is intended to help solve the task specified in the $<$TASK$>$ tags.
You must assist with improving the $<$PROMPT$>$ to solve the problem given by $<$TASK$>$.
A series of inputs and outputs, and evaluations are provided in $<$INPUT$>$ and $<$OUTPUT$>$.
After thinking about how to improve the prompt to improve the quality and accuracy of the outputs, output a single new prompt surrounded by $<$UPDATED-PROMPT$>$ tags.
\end{displayquote}

\subsection{Prompt + Output + Evaluation (critic)}

\label{appendix:prompts-indirect}

When using critic-based training, the generator's prompt optimization routine is unchanged. 
However, we use slightly different prompts for the critic's APO implementation, to reflect that the LLM must evaluate the evaluation, rather than the generator.

Given the evaluation $\evaluationShort$---in this case, the evaluation is scoring the critic's accuracy, rather than being an evaluation of the generator's output---the following prompt produces output feedback $\feedbackForOutput$:

\begin{displayquote}
    \small
    \obeylines
    You are a system for improving evaluator outputs based on feedback, part of a larger optimization system.
You will be provided with a set of examples as $<$INPUT$>$ $<$OUTPUT$>$ pairs.
Each pair was evaluated by an evaluator, with the instructions from the $<$EVALUATOR-PROMPT$>$ tag.
The evaluator's task is given in the $<$EVALUATOR-TASK$>$ tag.
The evaluator gave an evaluation in the $<$EVALUATION$>$ tag.
Feedback about the evaluator's evaluation is provided in the $<$EVALUATION-META-FEEDBACK$>$ tag.
Your task is to provide guidance on how to improve the evaluation based on the feedback.
Do not update the evaluator, and do not speculate on how to improve the evaluator. Only discuss how to improve the example evaluations.
After careful thought, provide feedback for each example evaluation in $<$EVALUATION-IMPROVEMENT$>$ tags.
\end{displayquote}

The next prompt produces prompt feedback $\feedbackForPrompt$:

\begin{displayquote}
    \small
    \obeylines
    You are a system for suggesting improvements to an evaluator prompt (set of instructions) based on feedback, part of a larger optimization system.
You will be provided with a set of examples as $<$INPUT$>$ $<$OUTPUT$>$ pairs.
These examples were evaluated by an evaluator, with the instructions from the $<$EVALUATOR-PROMPT$>$ tag.
The evaluator's task is given in the $<$EVALUATOR-TASK$>$ tag.
Suggestions to improve each evaluation are provided in the $<$EVALUATION-IMPROVEMENT$>$ tag.
Your task is to provide guidance on how to improve the evaluator prompt based on the evaluation feedback.
Do not update the evaluator; only provide feedback about how to improve the evaluator prompt based on the $<$EVALUATION-FEEDBACK$>$.
The feedback must help the evaluator arrive at the correct evaluation on its own.
The evaluator does not have access to any reference materials or code execution environments---it must reason on its own to arrive at a conclusion about the output.
After careful thought, provide feedback for the evaluator prompt in $<$EVALUATOR-PROMPT-FEEDBACK$>$ tags.
\end{displayquote}

Finally, the last prompt creates an updated prompt $\prompt'$:

\begin{displayquote}
    \small
    \obeylines
    You are a system for improving an evaluator prompt (set of instructions) based on feedback, part of a larger optimization system.
The current evaluator prompt is provided in $<$EVALUATOR-PROMPT$>$, and it is meant to solve the task given in the $<$EVALUATOR-TASK$>$ tag.
Feedback about how to improve the evaluator prompt is in the $<$EVALUATOR-PROMPT-FEEDBACK$>$ tag.
Apply the feedback to the evaluator prompt to improve it.
After careful thought, output a single new evaluator prompt surrounded by $<$UPDATED-EVALUATOR-PROMPT$>$ tags.
\end{displayquote}

\end{document}